# Progressive Multimodal Interaction Network for Reliable Quantification of Fish Feeding Intensity in Aquaculture

*Shulong Zhang[a, d, f], Mingyuan Yao[a, d, f], Jiayin Zhao[a, d, f], Daoliang Li[a, b, c, d, e, f], Yingyi Chen[a, b, c, d, e, f], Haihua Wang[a, b, c, d, e, f]**

*a National Innovation Center for Digital Fishery, Beijing 100083, China*

*b Key Laboratory of Smart Farming Technologies for Aquatic Animal and Livestock, Ministry of Agriculture and Rural Affairs, Beijing 100083, China*

*c State Key Laboratory of Efficient Utilization of Agricultural Water Resources, Beijing 100083, China*

*d Beijing Engineering and Technology Research Center for Internet of Things in Agriculture, Beijing 100083, China*

*e Key Laboratory of Digital Fisheries of Shandong Province, Yantai 264670, China*

*f College of Information and Electrical Engineering, China Agricultural University, Beijing 100083, China*

* Corresponding author.

Email: zhangshulong@cau.edu.cn (Shulong Zhang); yaomingyuan@cau.edu.cn (Mingyuan Yao); zhaojiayin@cau.edu.cn (Jiayin Zhao); dliangl@cau.edu.cn (Daoliang Li); chenyingyi@cau.edu.cn (Yingyi Chen); wang_haihua@163.com (Haihua Wang)

**Abstract:** Accurate quantification of fish feeding intensity is crucial for precision feeding in aquaculture, as it directly affects feed utilization and farming efficiency. Although multimodal fusion has proven to be an effective solution, existing methods often overlook the inconsistencies in responses and decision conflicts between different modalities, thus limiting the reliability of the quantification results. To address this issue, this paper proposes a Progressive Multimodal Interaction Network (PMIN) that integrates image, audio, and water-wave data for fish feeding intensity quantification. Specifically, a unified feature extraction framework is first constructed to map inputs from different modalities into a structurally consistent feature space, thereby reducing representational discrepancies across modalities. Then, an auxiliary-modality reinforcement primary-modality mechanism is designed to facilitate the fusion of cross-modal information, which is achieved through channel aware recalibration and dual-stage attention interaction. Furthermore, a decision fusion strategy based on adaptive evidence reasoning is introduced to jointly model the confidence, reliability, and conflicts of modality-specific outputs, so as to improve the stability and robustness of the final judgment. Experiments are conducted on a multimodal fish feeding intensity dataset containing 7089 samples. The results show that PMIN has an accuracy of 96.76%, while maintaining relatively low parameter count and computational cost, and its overall performance outperforms both homogeneous and heterogeneous comparison models. Ablation studies, comparative experiments, and real-world application results further validate the effectiveness and superiority of the proposed method. It can provide reliable support for automated feeding monitoring and precise feeding decisions in smart aquaculture.

# 1 Introduction

Aquaculture plays a crucial role in satisfying the growing global demand for aquatic products and providing a sustainable food source (Boyd et al., 2022). According to statistics from the Food and Agriculture Organization of the United Nations, the combined global output of fisheries and aquaculture reaches 223.2 million tons, and aquaculture production exceeds capture fisheries for the first time (FAO, 2024). However, as the scale of aquaculture continues to expand, problems such as increasing management complexity, severe feed waste, and the resulting frequent disease outbreaks become more prominent(Garlock et al., 2020; Naylor et al., 2023). Related studies have shown that behavioral changes displayed by fish during feeding reflect their feeding desire (Assan et al., 2021; MacGregor et al., 2020; Syafalni et al., 2024), and further quantification of feeding intensity therefore helps determine whether feed is supplied in excess or in insufficient amounts. However, the assessment of fish feeding behavior in actual production still relies on manual observation and record keeping by farm staff. Although this approach is intuitive and easy to implement, it is time-consuming, labor-intensive, and prone to substantial error. Feed delivery also largely depends on the experience and habits of farm staff. Although automated feeding machines with fixed schedules and quantities can reduce labor costs, they lack the ability to adjust feeding dynamically according to the real-time feeding demand of the fish population, which easily leads to underfeeding or feed waste. Therefore, achieving real-time, objective, and accurate quantification of fish feeding intensity is of great significance for improving feed utilization efficiency and aquaculture management.

In recent years, the rapid development of artificial intelligence and intelligent sensing technologies has provided a new technical approach for fish feeding state analysis. The researchers mainly utilized technologies such as computer vision (Y. Wu et al., 2024; Z. Wang et al., 2025; M. Wu et al., 2026), acoustics (Du, Xu, et al., 2023; Zeng et al., 2023; Helberg et al., 2024; Iqbal et al., 2025), and sensors (Adegboye et al., 2020; Ma et al., 2024; K. Zhang et al., 2025) to conduct their research. Among them, computer vision has become an important technical approach for fish feeding intensity analysis because of its low cost, non-invasive nature, and intuitive representation. However, visual information is highly susceptible to illumination changes, water turbidity, reflection interference, and fish occlusion, which often leads to unstable extraction of key features such as body color, texture, and shape. In contrast, acoustic technology is not restricted by illumination conditions or water turbidity and therefore shows strong potential for monitoring fish feeding processes (Zhang et al., 2025). With the aid of hydrophones, acoustic features such as frequency, energy, and waveform can be acquired in real time. Nevertheless, acoustic signals are also vulnerable to water-flow noise, device self-interference, and non-feeding sound sources. Behavior monitoring methods based on devices such as accelerometers can reflect fish activity characteristics from a kinematic perspective, but invasive or individualized monitoring not only may affect fish welfare, but also makes it difficult for the observations to accurately represent group-level feeding behavior. Overall, single-modality methods often struggle to balance robustness, accuracy, and practicality in complex aquaculture environments.

Compared with single-modality methods, multimodal fusion can comprehensively exploit the complementarity among different information sources and usually provides better generalization ability and environmental adaptability (W. Li et al., 2024). Existing studies have achieved promising results in fish feeding state analysis by integrating images, audio, and other sensing information (Yang et al., 2024; Gu et al., 2025; Zhang et al., 2025), but several limitations remain. Existing methods still struggle to cover the multidimensional representation of fish feeding behavior through modality selection. Most studies mainly focus on modalities that directly describe behavior. Although visual and acoustic modalities can characterize fish behavior and feeding activity in a relatively intuitive manner, they are easily affected by factors such as illumination variation, water scattering, and environmental noise in complex aquaculture environments, which limits the stability of their representations. In contrast, information that reflects external environmental feedback to feeding activity remains underused. Although some studies introduce water quality indicators such as dissolved oxygen (Zhang et al., 2025), local feeding activity usually has only a weak influence on global environmental parameters, and the variation of such indicators is often limited under stable culture conditions, which restricts their practical value for fish feeding intensity quantification. In addition, most existing methods adopt channel concatenation, static weight assignment, or simple late fusion strategies at the feature fusion level (Du et al., 2024; Zheng et al., 2024; Gu et al., 2025). These methods essentially still treat different modalities as independent information units that can be directly combined, while overlooking differences in response intensity, representation pattern, and discriminative emphasis under the same feeding state. As shown in Figure 1, the splashes are more obvious in some areas of Figure 1(a), but the overall water surface ripples are weak; the water surface ripples are larger in Figure 1(b), reflecting stronger environmental disturbances, but the fish's feeding behavior is not visually prominent. As a result, they find it difficult to capture the complex relationships among modalities and their differentiated contributions to the discrimination task. Existing studies still pay insufficient attention to joint reasoning at the decision level. Most methods keep fusion at the data level or feature level and lack explicit modeling of output inconsistency, conflict, and reliability across modalities. Related studies show that introducing cross-modal confidence allocation and evidential reasoning mechanisms helps alleviate decision conflicts and improve overall evaluation performance (Z. Zhao et al., 2025).

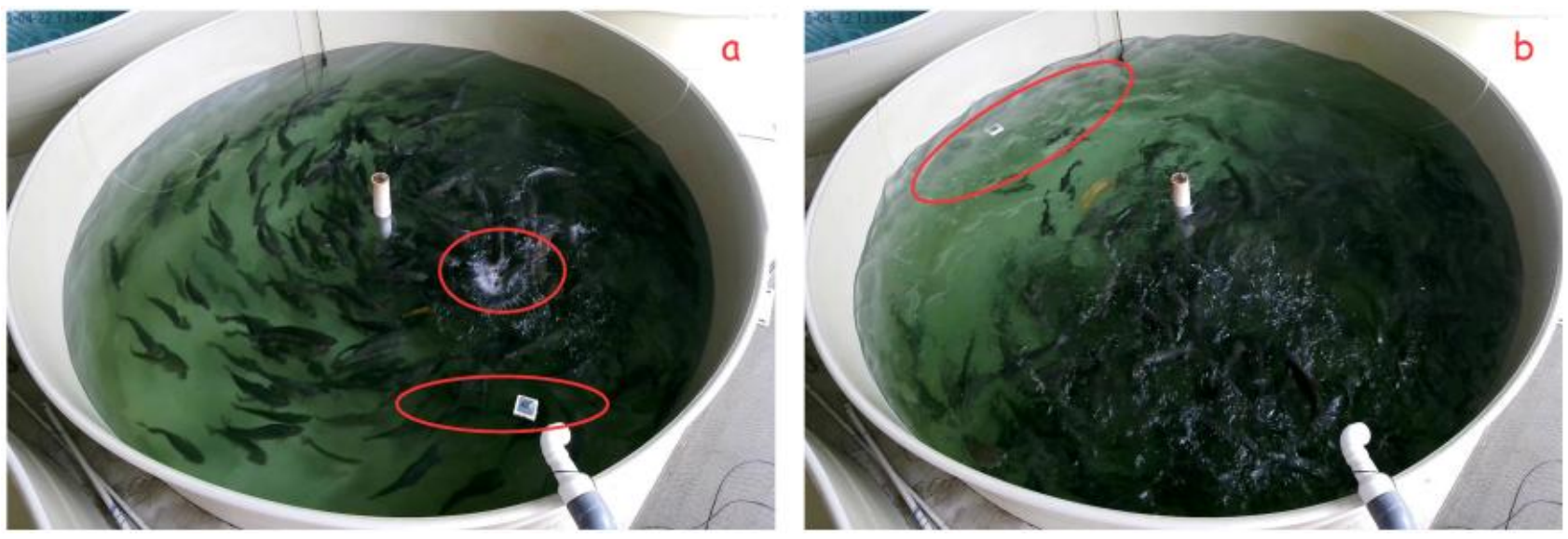

Figure 1. Inconsistent response of different modalities.

Therefore, to address the limited robustness of existing single-modality methods in complex scenarios and the remaining deficiencies of multimodal methods in modality selection, cross-modal interaction, and decision fusion for fish feeding intensity quantification, this paper proposes an innovative Progressive Multimodal Interaction Network (PMIN) to improve the accuracy and reliability of fish feeding intensity quantification. The main contributions of this study are as follows:

(1) A multimodal fish feeding intensity dataset that integrates image, audio, and water-wave data is constructed, which provides an experimental foundation for fish feeding state analysis by jointly covering direct behavioral representation and environmental response representation.

(2) A unified multimodal feature extraction framework is proposed. This framework employs the Universal Perception Large-Kernel ConvNet (UniRepLKNet) to encode feeding images, audio, and water-wave data into structurally consistent representations, thereby reducing representational discrepancies across modalities.

(3) A novel auxiliary-modality reinforcement primary-modality mechanism is proposed to progressively enhance the primary-modality representation through channel-aware recalibration and two-stage attention interaction. This design more effectively models the differentiated responses of different modalities under the same feeding state and improves the ability to mine complementary cross-modal information.

(4) A decision fusion strategy based on adaptive evidence reasoning is proposed to jointly model the confidence, reliability, and potential conflicts of modality-specific decisions, thereby producing more robust fish feeding intensity quantification results.

(5) Extensive experimental results show that the proposed PMIN significantly outperforms a variety of comparison methods on the fish feeding intensity quantification task, which verifies its effectiveness and superiority.

# 2 Related work

In the quantification of fish feeding intensity using computer vision techniques, Hu et al.(2015) analyzed the aggregation degree of the fish school during feeding and the area of the generated water splashes, and used the area ratio between the two as a feature parameter to characterize the hunger or satiety state of the fish school. Zhou et al.(2017) took the average perimeter of the Delaunay triangle as the aggregation index of the fish school to quantify the feeding intensity, and the results achieved a correlation coefficient of 0.945 with expert scores. However, the method was easily affected by factors such as water-surface reflection, fish crowding, and overlapping illumination. To address this issue, Hu et al.(2022) developed an intelligent fish farming system based on computer vision, which determined whether feeding should continue or stop by recognizing the magnitude of waves caused by fish feeding on pellets. Wu et al.(2024) proposed a new method that used feeding splash thumbnails to evaluate fish feeding intensity, which effectively removed the influence of water-surface reflection, bright spots, and ripples on the evaluation results. However, this method was not suitable for fry culture or low-density aquaculture environments, because splash phenomena generated by fish schools were not obvious in such scenarios. Zhang et al.(2024) jointly considered the dynamic changes in biomass, density, and feeding intensity of cultured fish, and proposed a recognition method based on dual labels and MobileViT-SENet, which showed excellent performance in fish feeding intensity recognition under different density conditions. In addition, to address the limited accuracy of lightweight models, Xu et al.(2024) integrated the convolutional block attention module and bi-directional long short-term memory into the lightweight evaluation network MobileViT, and the resulting feeding intensity evaluation model achieved a recognition accuracy of 98.61%. Zhao et al.(2024) proposed a new method for appetite assessment based on individual fish behavior. In their method, the ByteTrack model and a spatiotemporal graph convolutional neural network were used for individual fish tracking and motion feature extraction, respectively, thereby avoiding data loss caused by fish-school overlap. Although computer vision-based methods were convenient and effective, their resistance to interference remained insufficient in complex and diverse environments.

As an important carrier for the study of fish feeding behavior, acoustic information provides a scientific basis for feeding intensity quantification through its characteristic differences under different satiety states. Cao et al.(2021) obtained the feeding acoustic signals of largemouth bass in circulating aquaculture using passive acoustic techniques, and successfully filtered out the characteristic parameters that could measure the feeding activity from the mixed signals. Cui et al.(2022) further converted audio signals into Mel spectrogram (MS) features and used a Convolutional Neural Network (CNN) model to classify fish feeding intensity, achieving a mean average precision of 0.74. Although CNNs showed advantages in local receptive fields, they had limitations in capturing global features. To address this issue, Zeng et al.(2023) proposed an ASST model based on an attention mechanism, which achieved an accuracy of 96.16% in the task of fish feeding intensity quantification. In addition, Du et al.(2023a) extracted MS feature maps through multiple steps including preprocessing, fast Fourier transform, and Mel filter banks, and input them into the lightweight network MobileNetV3-SBSC for fish feeding intensity quantification. Although this method was computationally efficient, its accuracy was limited and it was not suitable for low-density aquaculture scenarios. Du et al.(2023b) further proposed a new method for fish feeding intensity quantification by combining MS, short-time Fourier transform feature maps, and constant-Q transform feature maps. Although this method achieved significantly higher accuracy than schemes using a single feature, the combination of multiple strategies increased model complexity. To address this problem, Iqbal et al.(2024) introduced a novel involutional neural network model, which could automatically capture labeled relationships and self-attention within the learned feature space, thereby yielding a lighter architecture and faster inference. Although acoustic technology is not constrained by illumination conditions or water turbidity and can display monitored data such as frequency, energy, and waveform in real time, it is highly susceptible to interference from non-feeding sounds. In high-density aquaculture environments, it is also necessary to pay attention to situations where fish collide with hydrophones.

In addition to computer vision and acoustic methods, sensors have also been applied to fish feeding intensity quantification. Biosensors are surgically inserted into the abdominal cavity or immobilized on the body surface to continuously monitor the fish behaviors and physiological parameters such as heart rate, temperature, orientation and acceleration over time (Brijs et al., 2021).

However, the invasiveness of implantable sensors poses a potential hazard to fish, limiting their practical application. Subakti et al.(2017) utilized sensors suspended on the water surface to sense the acceleration caused by the surface wave as a way to monitor the feeding activities of fish near surface water. Ma et al.(2024) introduced a six-axis inertial sensor to increase the data of angular velocity and angle, and proposed a time-domain and frequency-domain fusion model for quantifying the fish feeding intensity. The method can avoid the interference of equipment vibration noise, fish overlap, water turbidity and complex lighting. In addition, water quality parameters such as water temperature, dissolved oxygen and ammonia nitrogen compounds interact with the feeding behavior of fish (D. Li et al., 2020; K. Zhang et al., 2025). For example, the feeding activity of fish will lead to the localized decrease of dissolved oxygen concentration, and changes in dissolved oxygen concentration will directly affect fish appetite and food intake (D. Li et al., 2017). Zhao et al.(2019) took water temperature and dissolved oxygen concentration as input parameters of the adaptive neuro-fuzzy inference system model to determine fish feeding, and used a hybrid learning approach to optimize the parameters and fuzzy rule base. The Nash-Sutcliffe efficiency coefficient and root mean squared error of the model outperformed traditional fuzzy logic control and artificial feeding methods. Chen et al.(2020) proposed a fish intake prediction model based on back propagation neural network and mind evolutionary algorithm, which successfully established the mapping relationship between fish intake and environmental factors and biomass by using temperature, dissolved oxygen, weight and number of fish as input variables, avoiding the subjectivity of traditional methods.

The rapid development of multimodal fusion technology has also provided new ideas for quantifying fish feeding intensity. Syafalni et al.(2024) proposed a multimodal sensor-based method for fish appetite detection, which used residual (2+1)-dimensional CNN and dense network to process video and accelerometer data, with an accuracy rate of up to 99.09% on the validation set. Du et al.(2024) developed a multi-modal fusion framework called MFFFI, which combines deep features from audio, video and acoustic data and outperforms mainstream single modality methods. Hu et al.(2023) added a multimodal transfer module and adaptive weights to the MulT algorithm to achieve effective fusion of feature vectors and dynamic adjustment of modal contributions, and further optimized the number of cross-modal transformers. Xu et al.(2023) proposed a multi-level fusion model based on sound and visual features to identify fish swimming and feeding behaviors under complex conditions, which fuses modal features from different stages through a designed jump connection module. Yang et al.(2024) designed a U-shaped bilinear fusion structure to achieve more interaction between sound and visual features, and introduced a time aggregation and pooling layer to retain the optimal feature information of fish. In addition, Zheng et al.(2024) used near-infrared images and depth maps to characterize fish feeding behavior, combining the feature information of feeding dynamics, water level fluctuation and feeding audio by weighted fusion. Gu et al.(2025) developed an audio-video aggregation module consisting of self-attention and cross-attention mechanisms and introduced a lightweight separable convolutional feedforward module to reduce model complexity, achieving a balance between the speed and accuracy of quantifying fish feeding intensity.

# 3 Materials and methods

## 3.1 Dataset

### 3.1.1 Data collection

Data collection is conducted at the Guoyu Green Smart Aquaculture Facility in Shangluo, Shaanxi Province, China, and the experimental platform is shown in Figure 2. The entire experimental system consists of a recirculating aquaculture unit, an optical imaging unit, an underwater acoustic monitoring unit, and a water-surface fluctuation sensing unit. The aquaculture unit employs a standardized recirculating water system with a circular aquaculture pond of 4 m in diameter, and the water depth is maintained at 1±0.2 m. The experimental subjects are adult rainbow trout (Oncorhynchus mykiss) with an average body length of 35±5 cm and a weight of 1.35±0.15 kg. The optical imaging unit uses an industrial-grade 4K camera (3840×2160 resolution, 30 fps), which is mounted vertically at a height of 3 m above the aquaculture pond. The acoustic monitoring unit uses a high-frequency hydrophone (20 Hz-50 kHz) fixed at the center of the tank. The water-surface fluctuation sensing unit uses a six-axis accelerometer (WT9011DCL-BT50, 200 Hz), which is waterproofed and mounted on a floating platform on the water surface.

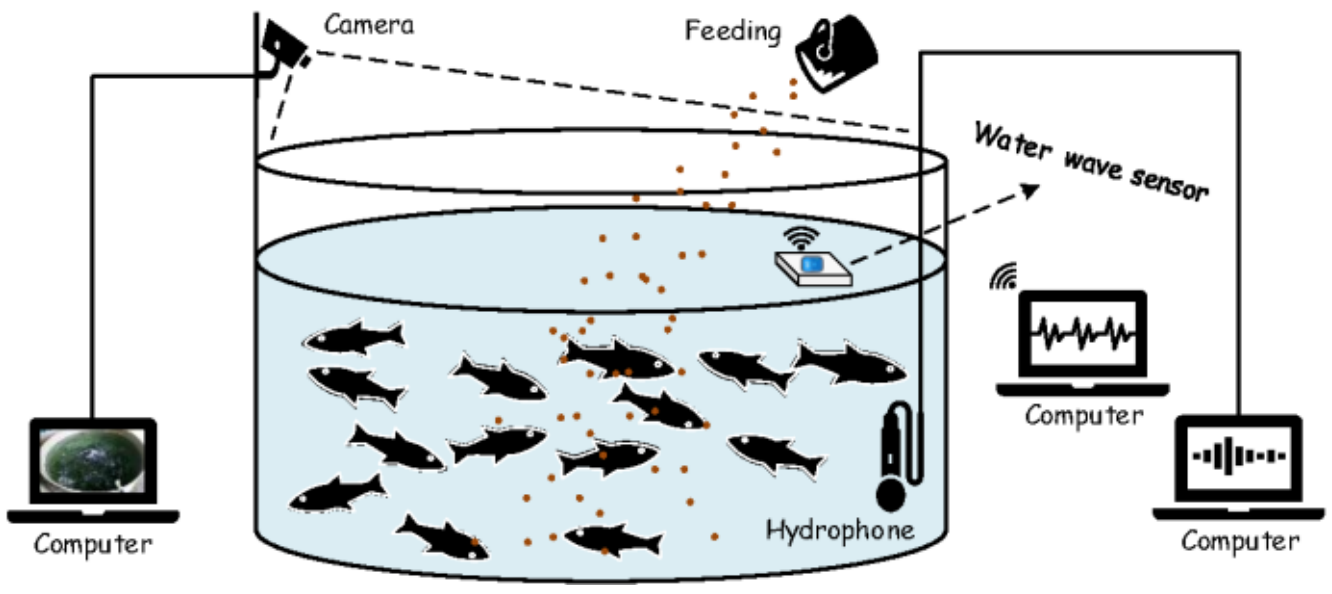

Figure 2. Data collection platform.

During the collection, the water temperature is maintained at 17±2°C, the dissolved oxygen concentration is kept at 12±2 mg/L, and the pH is controlled at 6.7±0.2. Feeding strictly follows the standardized operating procedures of the aquaculture facility and is conducted twice daily at 8:00 and 16:00. The feeding amount for each session is calculated as 1.5% of the total fish biomass (Azim & Little, 2008). Each feeding session is divided into three rounds, with each round lasting 3 min and a 1 min buffer interval inserted between adjacent rounds. The feed amount decreases across the three rounds at a ratio of 4:3:3. During feeding, the feeding area is dynamically adjusted through manual observation to ensure uniform feed dispersion. Image, audio, and water-wave data are synchronously collected during the feeding process to ensure temporal consistency across modalities.

### 3.1.2 Data processing

To ensure the consistency of multimodal samples, this study performs millisecond-level temporal alignment of video, audio, and water-wave data so that different modalities can accurately correspond to the feeding state of the fish school within the same time window. The continuous data are then segmented and sampled using a sliding window with a fixed width of 1 second and an overlap ratio of 50%, thereby preserving the dynamic continuity of feeding behavior over short time scales. Each sample consists of three synchronously acquired modalities, including a single-frame RGB image, 1 second of audio, and time-series data recorded by the sensor along the X/Y/Z axes, specifically including acceleration, angular velocity, and Euler angle measurements.

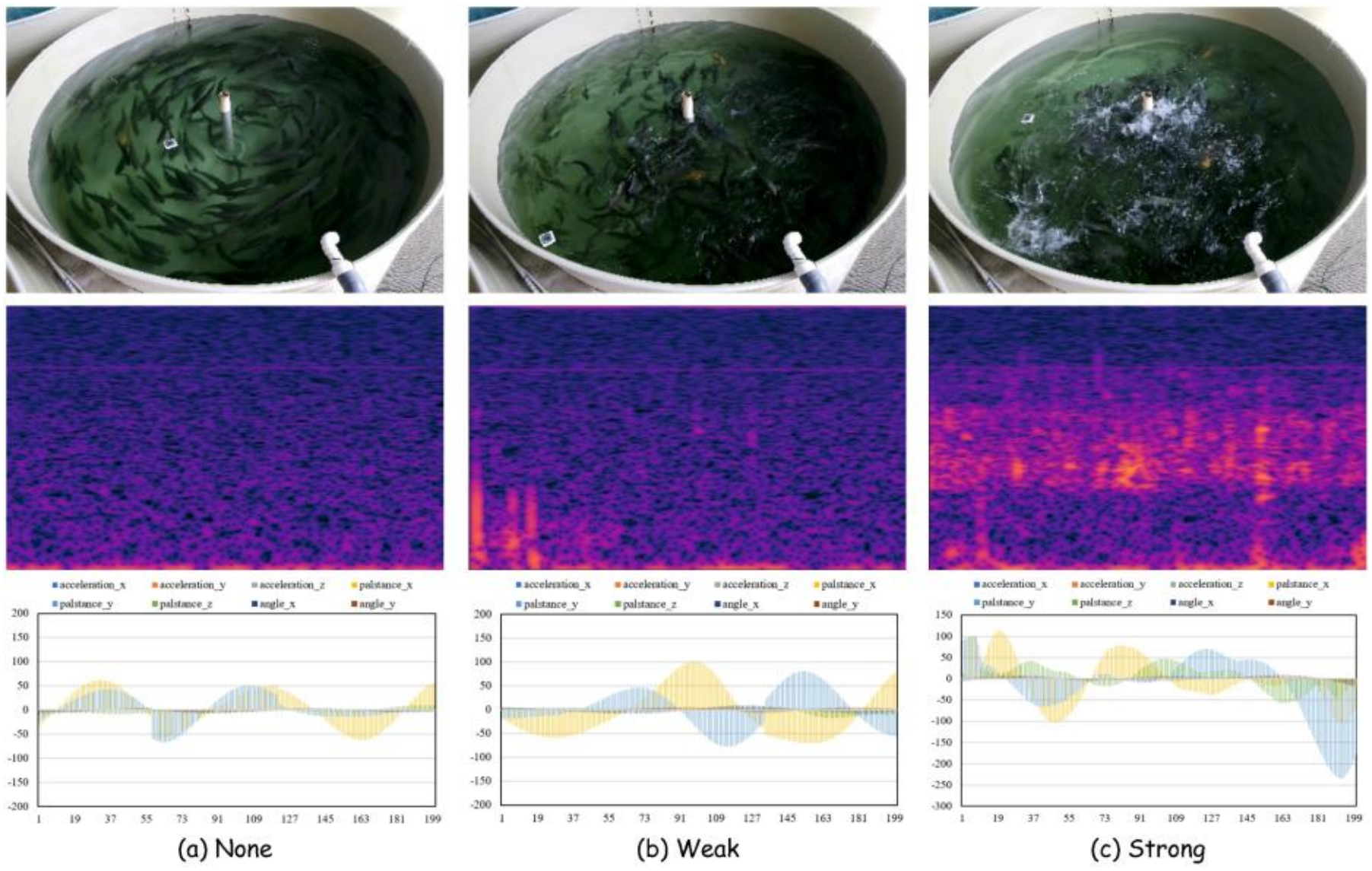

Figure 3. Multimodal data visualization of different feeding intensities.

By considering the practical aquaculture environment, the fish feeding habits of rainbow trout, expert knowledge from aquaculture practice, and existing feeding intensity quantification standards (Øverli et al., 2006), this study categorizes feeding intensity into three levels: strong, weak, and none. Strong feeding intensity corresponds to high-density aggregation of the fish school and pronounced splashing and turbulent disturbances caused by vigorous feed competition. Weak feeding intensity is characterized by scattered foraging behavior accompanied by local splash disturbances. In contrast, the non-feeding state is characterized by regular swimming activity and a relatively calm water surface. Figure 3 shows typical multimodal samples under different feeding intensity categories, including images, Mel spectrograms of the corresponding audio signals (Kong et al., 2020), and feature distributions of water-wave data. After rigorous screening and manual annotation, a multimodal fish feeding intensity dataset containing 7089 temporally synchronized samples is finally constructed. The dataset is divided into the training set, validation set, and test set at a ratio of 8:1:1, as shown in Table 1.

Table 1. Distribution of fish feeding intensity datasets.

| Feeding intensity | Train | Validation | Test | Total |
| --- | --- | --- | --- | --- |
| Strong | 1927 | 241 | 241 | 2409 |
| Weak | 1881 | 236 | 236 | 2353 |
| None | 1861 | 233 | 233 | 2327 |
| Total | 5669 | 710 | 710 | 7089 |

### 3.2 Proposed method

This study proposes a novel Progressive Multimodal Interaction Network (PMIN) for fish feeding intensity quantification, and the overall architecture is shown in Figure 4. It mainly consists of three stages: unified feature extraction, progressive multimodal feature interaction, and decision fusion. Specifically, in the unified feature extraction stage, UniRepLKNet is adopted as the backbone to encode the input image, audio, and water-wave data into structurally consistent feature representations. This design reduces representational discrepancies across modalities and establishes a unified feature basis for subsequent cross-modal interaction. Next, corresponding interaction branches are constructed for the three modalities. In each branch, the current modality serves as the primary modality, while the other two modalities serve as auxiliary modalities. Through the Auxiliary-modality Reinforcement Primary-modality Mechanism (ARPM), auxiliary information is progressively introduced to enhance the representation of the primary modality. This mechanism more effectively captures the differentiated responses of different modalities under the same feeding state and improves the collaborative representation capability of multimodal features. Finally, a decision fusion method based on Adaptive Evidence Reasoning (AER) is used to jointly model the outputs of all modality branches. By comprehensively considering the confidence, reliability and potential conflicts of each branch, a more reliable quantitative result of fish feeding intensity can be obtained.

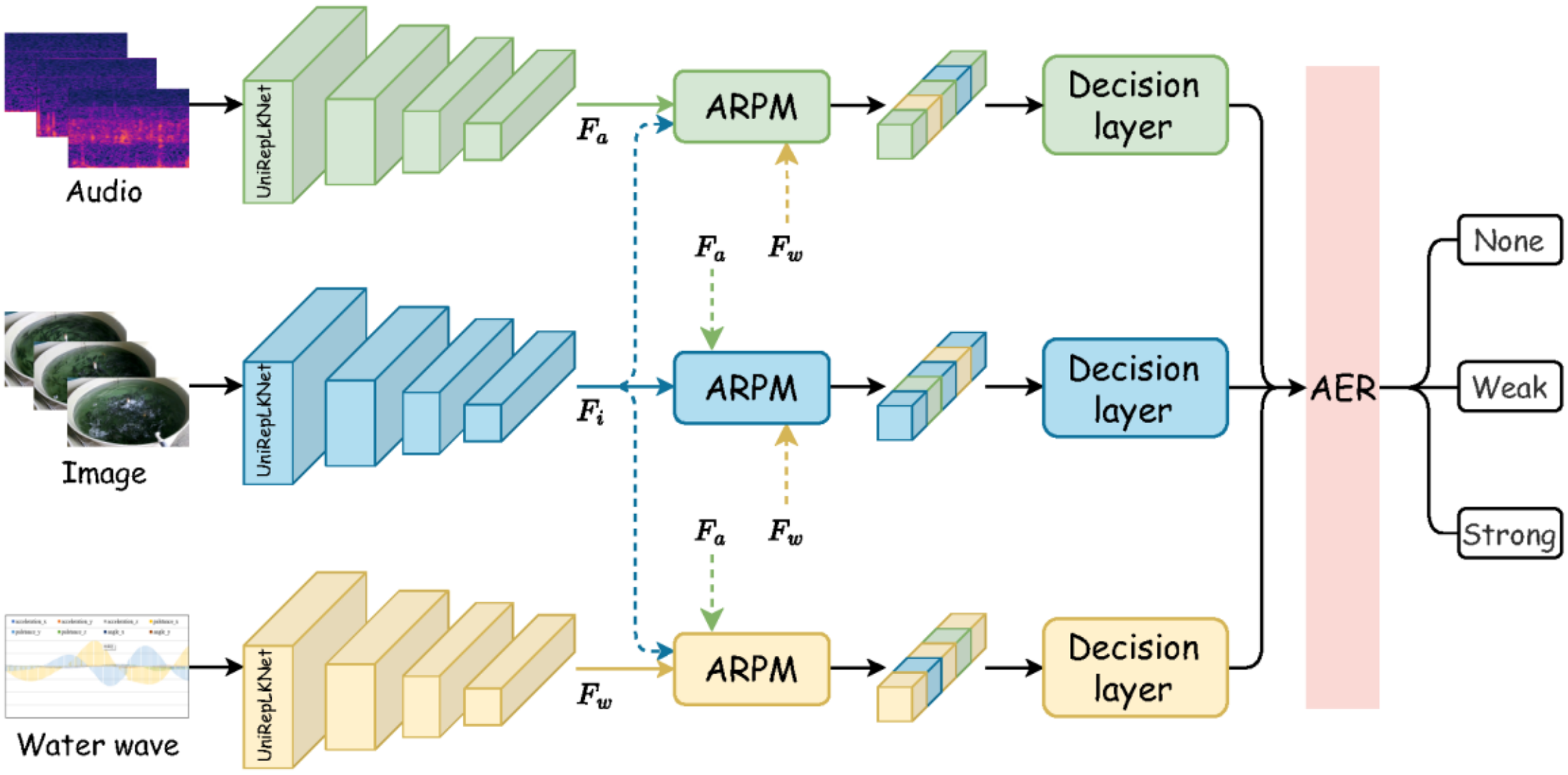

Figure 4. Overall architecture of the PMIN.

#### 3.2.1 Multimodal unified feature extraction framework

In fish feeding intensity quantification, existing multimodal methods usually adopt modality-specific heterogeneous encoders to extract features from different types of data independently. Although this strategy allows dedicated network architectures to be designed for the characteristics of each modality, it also tends to introduce cross-modal representational inconsistency. Because the features produced by different encoders often differ in embedding space, scale distribution, and statistical properties, additional mapping or adaptation modules are usually required to achieve feature alignment and cross-modal interaction. To address this issue, this study designs a unified multimodal feature extraction framework that adopts UniRepLKNet as the backbone to encode image, audio, and water wave data into structurally consistent feature representations. By extracting multimodal features in a shared representation space, the proposed model effectively alleviates representational discrepancies among modalities and provides a unified and stable feature basis for subsequent multimodal interaction. The architecture of UniRepLKNet is shown in Figure 5.

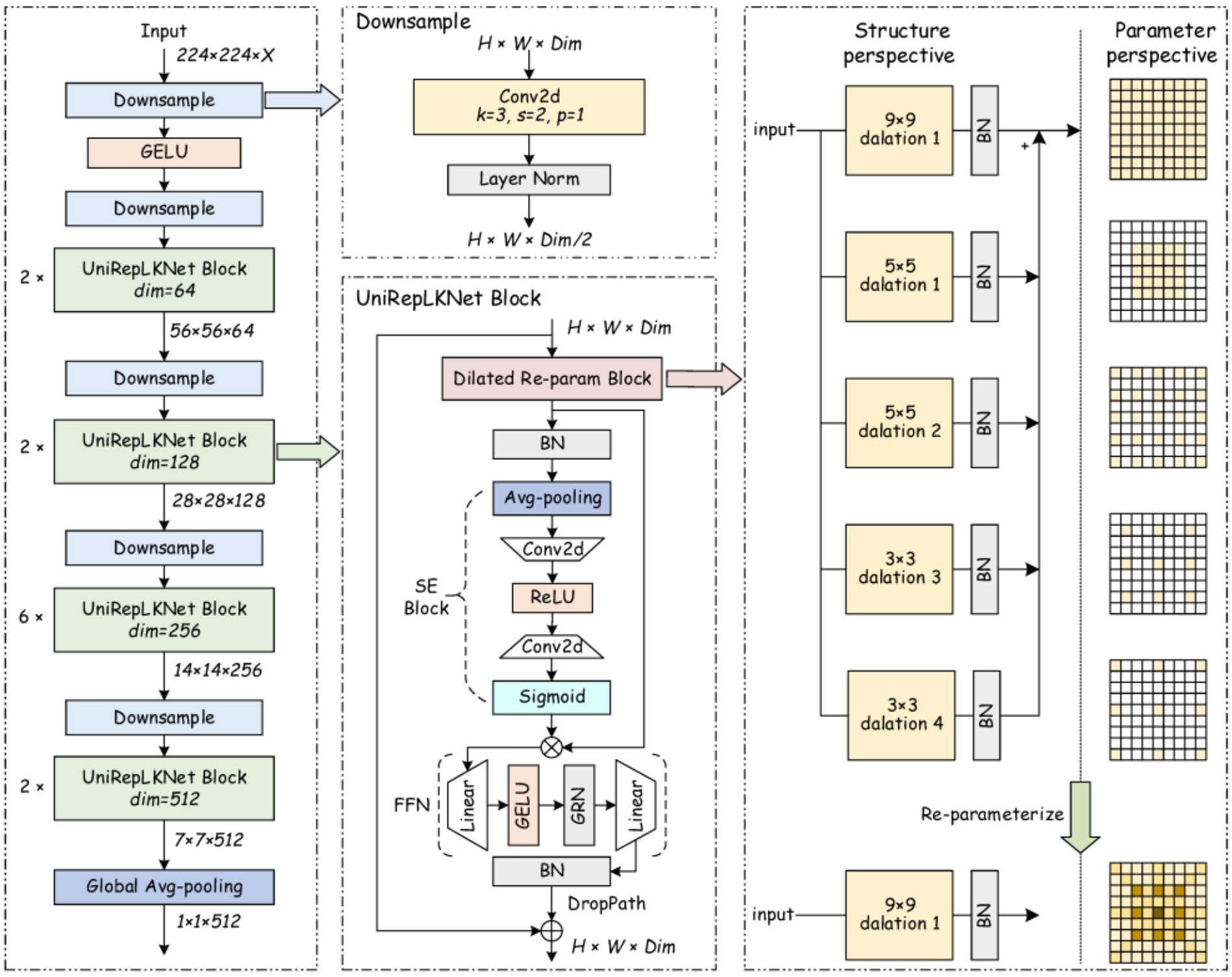

Figure 5. The structure of UniRepLKNet.

This study adopts UniRepLKNet as the unified backbone mainly because it is not only designed for image recognition, but also demonstrates strong generalization ability across a variety of tasks, including audio, point cloud, and time series analysis (Ding et al., 2024). The effectiveness of UniRepLKNet mainly stems from several key designs in large-kernel convolutional networks. UniRepLKNet introduces efficient components such as Squeeze-and-Excitation (SE) and bottleneck modules into its local structure, which increase network depth while controlling computational cost, thereby improving the modeling capability for complex features. The proposed Dilated Reparam Block can be equivalently transformed into large-kernel convolution through structural re-parameterization, which enhances the model's ability to perceive sparse spatial patterns and complex structures. In addition, the selection of kernel size is matched to the specific task and network architecture. Although introducing an excessively large receptive field in the shallow stages may affect low-level feature learning, large convolution kernels do not inherently weaken the representation ability of the model. Considering the characteristics of the present task and following related work (Ding et al., 2022), this study adopts a 13×13 convolution kernel to balance contextual modeling capability and computational cost. As the network becomes deeper, more large-kernel layers are further replaced with depthwise 3×3 convolution blocks, thereby improving feature abstraction capability while maintaining computational efficiency.

In multimodal data processing, UniRepLKNet provides a unified and flexible implementation scheme. For non-image modalities, the main network structure does not need to be modified, and only the input needs to be converted into an embedding map in the C×H×W format. Accordingly, in this study, the image data are directly represented as a 3×224×224 tensor after resizing. The audio signal is collected in a dual-channel format, and the two channels are separately converted into Mel spectrograms (Kong et al., 2020), which are then represented as a 2×224×224 tensor after adaptive average pooling. For the water-wave data, this study follows the simplified processing strategy of UniRepLKNet (Ding et al., 2024). The raw time-series signal is first mapped into a latent tensor space and then reshaped into a single-channel two-dimensional representation. After dimension adjustment, a 1×224×224 input tensor is obtained. Through these transformations, all three modalities can be encoded within a unified network architecture while maintaining consistency in feature representation format.

### 3.2.2 Auxiliary-modality reinforcement primary-modality mechanism

Considering that different modalities often exhibit different response strengths and discriminative emphases under the same feeding state, this study further proposes an Auxiliary-modality Reinforcement Primary-modality Mechanism (ARPM) to fully exploit complementary information across modalities. The overall structure of ARPM is shown in Figure 6.

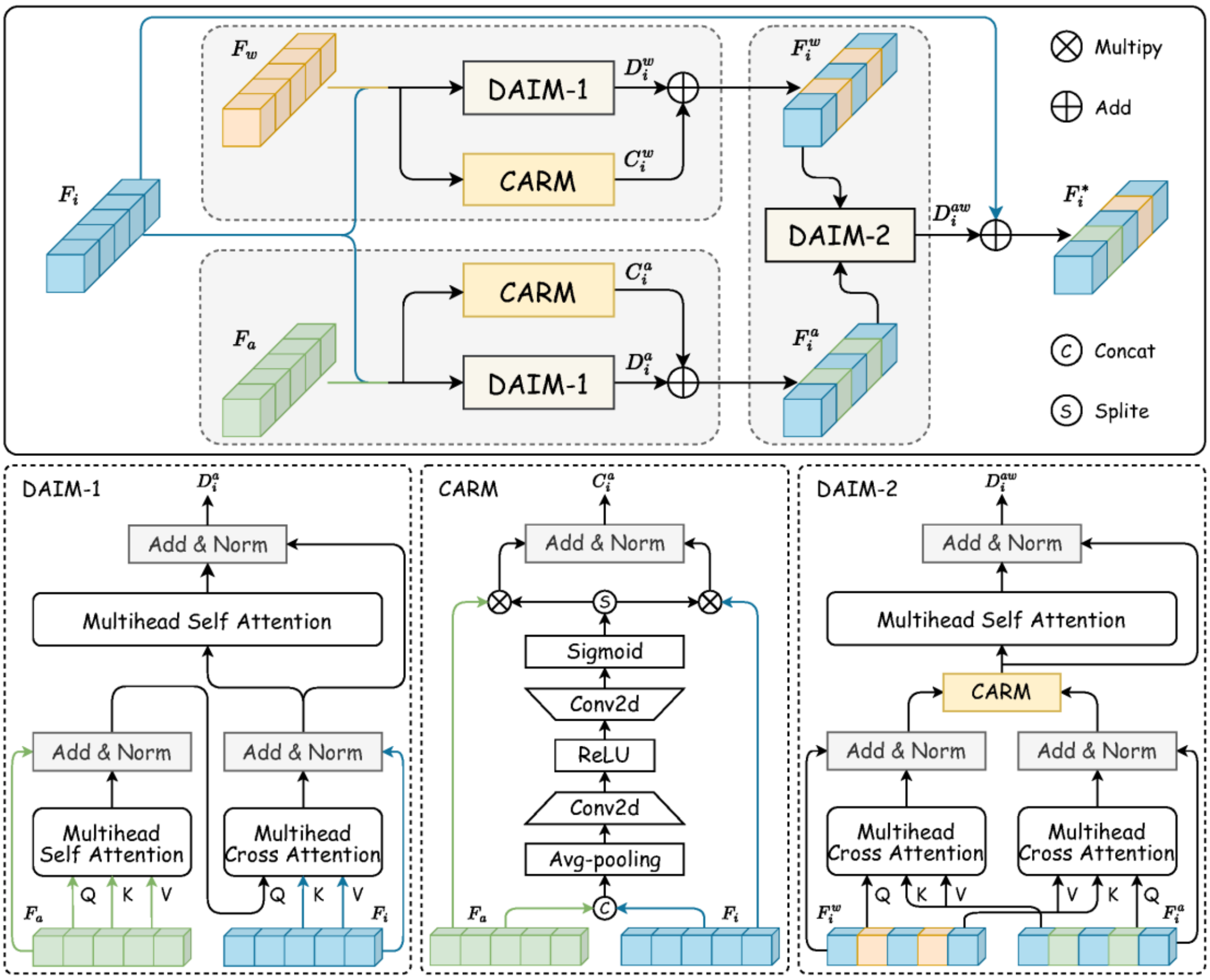

Figure 6. The structure of ARPM.

The core idea of ARPM is to progressively enhance the representation of the primary modality by gradually introducing effective information from auxiliary modalities. To this end, ARPM constructs a two-stage progressive interaction architecture, which mainly consists of a Channel Aware Recalibration Module (CARM) and a Dual-mode Attention Interaction Module (DAIM). CARM adaptively recalibrates multimodal features at the channel level to mitigate inter-modal differences, while DAIM models the interaction between the primary and auxiliary modalities. To accommodate the modeling requirements of features at different levels, DAIM is further designed into two functionally complementary variants, namely DAIM-1 and DAIM-2, which are used for shallow interaction modeling and deep semantic fusion, respectively.

In the shallow interaction stage, ARPM constructs two enhancement paths centered on the primary modality feature $F_i$, which fuse the auxiliary modality features $F_a$ and $F_w$, respectively, to achieve feature enhancement. At this stage, DAIM-1 performs initial fusion of the original input features through a serial combination of Multi-head Self-Attention (MHSA) and Multi-head Cross-Attention (MHCA). Specifically, MHSA is first applied to model the internal correlations within the auxiliary modality features, so that informative components are preserved while redundant information is suppressed (Vaswani et al., 2017). Then, MHCA is used with the primary modality feature as the query and the auxiliary modality feature as the key and value, so that discriminative information in the auxiliary modality is selectively injected into the primary modality representation, thereby achieving cross-modal alignment and complementary enhancement(Y. Wang, Li, et al., 2023a). Finally, MHSA is applied again to aggregate the interacted features, yielding the preliminarily enhanced features $D_i^a$ and $D_i^w$.The calculation process for MHSA is as follows:

For the auxiliary modality feature $F_a$, the query, key, and value matrices are first generated through linear projection:

$$Q_a = F_a \mathbf{W}^Q,\ K_a = F_a \mathbf{W}^K,\ V_a = F_a \mathbf{W}^V \tag{1}$$

Where $\mathbf{W}^Q$, $\mathbf{W}^K$ and $\mathbf{W}^V$ denote the learnable projection matrices for the query, key, and value, respectively. The output of the $h$-th self-attention head is expressed as:

$$\text{head}_a^h = \text{Attention}\left(Q_a^h, K_a^h, V_a^h\right) = \text{Softmax}\left(\frac{Q_a^h \left(K_a^h\right)^\top}{\sqrt{d_h}}\right) V_a^h \tag{2}$$

Where $d_h$ denotes the feature dimension of a single attention head; $\top$ denotes the transpose operation. The outputs of all attention heads are concatenated and linearly projected to obtain the output:

$$\text{MHSA}\left(F_a\right) = \text{Concat}\left(\text{head}_a^1, \ldots, \text{head}_a^H\right) \mathbf{W}_s^O \tag{3}$$

Where $H$ denotes the number of attention heads; $\mathbf{W}_s^O$ is the output projection matrix of MHSA. This mechanism effectively models the long-range dependency among different positions within a single modality, thereby enhancing intra-modal semantic consistency and global contextual representation capability.

Unlike MHSA, MHCA's query vector comes from the auxiliary modality feature $F_a$, while its key and value vectors come from the primary modality feature $F_i$. That is:

$$Q_a = F_a \mathbf{W}^Q,\ K_i = F_i \mathbf{W}^K,\ V_i = F_i \mathbf{W}^V \tag{4}$$

Similarly, the output of the $h$-th cross-attention head can be expressed as:

$$\text{head}_{ia}^h = \text{Attention}\left(Q_a^h, K_i^h, V_i^h\right) = \text{Softmax}\left(\frac{Q_a^h\left(K_i^h\right)^\top}{\sqrt{d_h}}\right)V_i^h \tag{5}$$

Finally, the output of MHCA is expressed as:

$$\text{MHCA}\left(F_i, F_a\right) = \text{Concat}\left(\text{head}_{ia}^1, \ldots, \text{head}_{ia}^H\right)\mathbf{W}_c^O \tag{6}$$

Meanwhile, to alleviate the modality weight solidification problem that may be caused by the distinction between primary and auxiliary modalities (Wang et al., 2023), ARPM introduces CARM into both branches to achieve adaptive recalibration of cross-modal features. CARM can be viewed as an extension of SENet (J. Hu et al., 2018). CARM first aggregates information from different modalities through a concatenation operation, and then sequentially performs squeeze and excitation operations to generate channel-wise attention weights. The resulting weight vector is then split and applied to the corresponding modality branches, thereby completing dynamic modeling and adaptive recalibration of channel importance. Through this process, CARM alleviates representational discrepancies between the primary and auxiliary modalities and improves the consistency and discriminative capability of the fused features. Finally, the outputs of DAIM-1 and CARM are fused by element-wise addition to obtain two shallow enhanced features $F_i^a$ and $F_i^w$.

In the deep interaction stage, ARPM further employs DAIM-2 to fuse the shallow enhanced features. Unlike the serial attention structure in DAIM-1, DAIM-2 adopts a parallel cross-modal attention mechanism to extract complementary information from different directions, which can be formulated as:

$$\tilde{F}_i^a = \text{MHCA}\left(F_i^a, F_i^w\right),\quad \tilde{F}_i^w = \text{MHCA}\left(F_i^w, F_i^a\right) \tag{7}$$

Where $\tilde{F}_i^a$ denotes the enhanced feature obtained after interaction with $F_i^w$ as the query and $F_i^a$ as the key and value; $\tilde{F}_i^w$ denotes the feature representation obtained from the reverse interaction. This bidirectional parallel interaction scheme captures deep complementary semantics from different directions, allowing the primary modality to receive multi-level information from different auxiliary modalities in the deep representation space, thereby enhancing the discriminative capability of deep features.

Meanwhile, to further alleviate the semantic inconsistency that may arise during cross-modal interaction (Wang et al., 2023), CARM is introduced again to perform adaptive fusion and recalibration of the interaction results. The fused feature is then fed into an MHSA module to model the global dependency in the deep representation space, yielding the deep fusion feature:

$$D_i^{aw}=\mathrm{MHSA}\left(\mathrm{CARM}\left(\tilde{F}_i^{a},\tilde{F}_i^{w}\right)\right) \tag{8}$$

Finally, the deep fusion feature is residually fused with the original primary modality feature $F_i$ to produce the final enhanced primary modality representation:

$$F_i^{*}=F_i+D_i^{aw} \tag{9}$$

Through the above progressive interaction process, ARPM gradually absorbs discriminative cues from auxiliary modalities while preserving the core semantic information of the primary modality. Specifically, MHSA strengthens the model's ability to capture global dependency within each modality, MHCA improves semantic alignment and information complementarity across modalities, and CARM further alleviates cross-modal representational discrepancies. Together, these components form a stable multimodal feature enhancement framework and provide effective representational support for fish feeding intensity quantification.

### 3.2.3 Decision fusion based on adaptive evidence reasoning

To address the decision conflicts and confidence inconsistency that may arise across different modalities under the same feeding state, this study designs a decision fusion strategy based on Adaptive Evidential Reasoning (AER). First, the enhanced features obtained from multimodal feature interaction are input into the corresponding decision networks composed of fully connected layers to generate classification confidence results. On this basis, the output of each branch is transformed into a basic probability assignment, and multi-source evidence is fused according to the AER (Yang et al., 2018; X. Xu et al., 2020). This strategy jointly models the conflict relationships, reliability differences, and uncertainty information among decisions from different branches. As a result, it preserves effective discriminative evidence while suppressing the interference caused by conflicting evidence, and finally produces a more robust global decision output. The calculation process for AER is as follows:

1) The independent classification result of each modality is regarded as one piece of evidence in the rule. Assume that there are $M$ independent pieces of evidence, where $M$ denotes the number of modalities, and the $m$-th piece of evidence is denoted by $e_m$ $(m=1,2,\ldots,M)$.

2) Each piece of evidence is transformed into the form of a confidence distribution. The $m$-th piece of evidence is expressed as:

$$e_m = \left\{ \left( \theta_n, p_{n,m} \right), \forall \theta_n \subseteq \Theta \right\} \tag{10}$$

where $\theta_n$ denotes a feeding intensity category; $p_{n,m}$ represents the confidence assigned by the $m$ -th piece of evidence to category $\theta_n$ , satisfying $\sum_{n=1}^{N} p_{n,m} = 1$ ; $\Theta = \{\theta_1, \theta_2, \cdots, \theta_N\}$ denotes the frame of discernment; $p_{\Theta,m}$ denotes global ignorance.

3) In the evidential reasoning rule, the evidence weight $w_m$ is used to describe the preference degree of the decision maker for a certain evidence item, while the evidence reliability $r_m$ is used to characterize the credibility of the evidence source. Traditional methods typically pre-determine these parameters, but this introduces a significant degree of subjectivity (Jiang et al., 2023). Therefore, this study treats $w_m$ and $r_m$ as learnable parameters and adaptively optimizes them during training. The weighted confidence distribution after introducing reliability is expressed as:

$$\tilde{e}_m = \left\{ \left( \theta, \tilde{p}_{\theta,m} \right), \forall \theta \subseteq \Theta, \left( P(\Theta), \tilde{p}_{P(\Theta),m} \right) \right\} \tag{11}$$

$$\tilde{p}_{\theta,m} = \begin{cases} 0 & , \ \theta = \varnothing \\ c_{rw,m} p_{\theta,m} & , \ \theta \subseteq \Theta, \theta = \varnothing \\ c_{rw,m} \left( 1 - r_m \right) & , \ \theta = P(\Theta) \end{cases} \tag{12}$$

$$\sum_{\theta \subseteq \Theta} \tilde{p}_{\theta,m} + \tilde{p}_{P(\theta),m} = 1 \tag{13}$$

Where $P(\Theta)$ denotes the power set; $p_{\theta,m} = w_m p_{\theta,m}$ ; $c_{rw,m} = 1/1 + w_m - r_m$ is the normalization factor; $\varnothing$ denotes the empty set.

4) by performing fusion equation (14) the combined belief degree that the $M$ pieces of evidence assign to feeding sample $x$ belonging to intensity level $\theta_n$ is obtained as:

$$P_{\theta_n}(x) = \frac{L \left[ \prod_{m=1}^{M} c_{rw,m} \left( 1 - r_m + \alpha_{\theta_n,m} \right) - \prod_{m=1}^{M} c_{rw,m} \left( 1 - r_m \right) \right]}{1 - L \prod_{m=1}^{K} c_{rw,m} \left( 1 - r_m \right)} \tag{14}$$

Where $L$ denotes the normalization factor, which is calculated as follows:

$$L = \left[ \sum_{n=1}^{N} \left( \prod_{m=1}^{M} c_{rw,m} \left( 1 - r_m + \alpha_{\theta_n,m} \right) \right) - \left( N - 1 \right) \left( \prod_{m=1}^{M} c_{rw,m} \left( 1 - r_m \right) \right) \right]^{-1} \tag{15}$$

5) Through the above computation, the combined belief degrees of sample $x$ over all intensity levels are obtained as $\left( P_{\theta_1}(x), P_{\theta_2}(x), \cdots, P_{\theta_N}(x) \right)$. The intensity level $\theta_n$ corresponding to the maximum belief degree is taken as the final quantification result of the sample.

## 3.3 Model evaluation metrics

This study uses the Accuracy, Precision, Recall and F1-Score derived from the confusion matrix to evaluate the performance of the model. Accuracy serves as a fundamental evaluation metric for multiclass classification tasks and measures the overall proportion of correctly classified samples across all categories. Precision reflects the reliability of samples predicted as the positive class by the model. Recall measures the model's ability to identify positive samples. F1-Score is the harmonic mean of Precision and Recall and is used to provide a comprehensive evaluation of model performance. The specific formulations are given as follows:

$$\text{Accuracy} = \frac{\text{TP} + \text{TN}}{\text{TP} + \text{TN} + \text{FP} + \text{FN}} \tag{16}$$

$$\text{Precision} = \frac{\text{TP}}{\text{TP} + \text{FP}} \tag{17}$$

$$\text{Recall} = \frac{\text{TP}}{\text{TP} + \text{FN}} \tag{18}$$

$$\text{F1-Score} = \frac{2 \times \text{Precision} \times \text{Recll}}{\text{Precision} + \text{Recll}} \tag{19}$$

Where $\text{TP}$ and $\text{FN}$ denote the number of samples in which the actual positive category is predicted to be positive and negative, respectively; $\text{TN}$ and $\text{FP}$ denote the number of samples in which the actual negative category is predicted to be negative and positive, respectively. In addition, this study also uses Parameters (Params) and Floating-Point Operations Per second (FLOP) to evaluate the total number of parameters and computational complexity of the model.

# 4 Experiments and results

## 4.1 Implementation details

The experiments were conducted on a workstation equipped with an Intel Core i9-13900K CPU (32 threads), 128 GB RAM, and an NVIDIA GeForce RTX 4090 GPU. The operating system is Ubuntu 23.04, and all models were implemented using the PyTorch deep learning framework. The training settings are shown in Table 2. All compared models were trained under the same experimental settings to ensure a fair comparison.

Table 2. Training settings for the model.

| Configuration | Parameter | Configuration | Parameter |
|---|---|---|---|
| Batch size | 64 | Learning rate scheduler | ReduceLROnPlateau |
| Training epoch | 100 | Patience | 5 |
| Optimizer | Adam | Attenuation coefficient | 0.5 |
| Initial learning rate | 1e-3 | Random seed | 42 |

## 4.2 Model training process

Figure 7 shows the curves of accuracy and loss for PMIN on the training and validation sets. In the early stage of training, both the training loss and validation loss decrease rapidly, while the training accuracy and validation accuracy increase synchronously and reach relatively high levels within a small number of epochs. This indicates that the model can effectively extract multimodal features related to fish feeding intensity discrimination at the early stage of optimization and exhibits a fast convergence rate. As training proceeds, the validation accuracy remains consistently close to the training accuracy, suggesting that the learned feature representations have good generalization ability. Meanwhile, the training loss gradually approaches zero, whereas the validation loss remains at a low level with only slight fluctuations, indicating that no obvious overfitting occurs during training and that the overall optimization process is stable. In the later stage of training, both the training accuracy and validation accuracy curves become gradually flat, and the loss curves fluctuate only within a narrow range, which suggests that the parameter updates become stable and the model progressively converges. Overall, PMIN demonstrates fast convergence, good stability, and strong generalization performance during training.

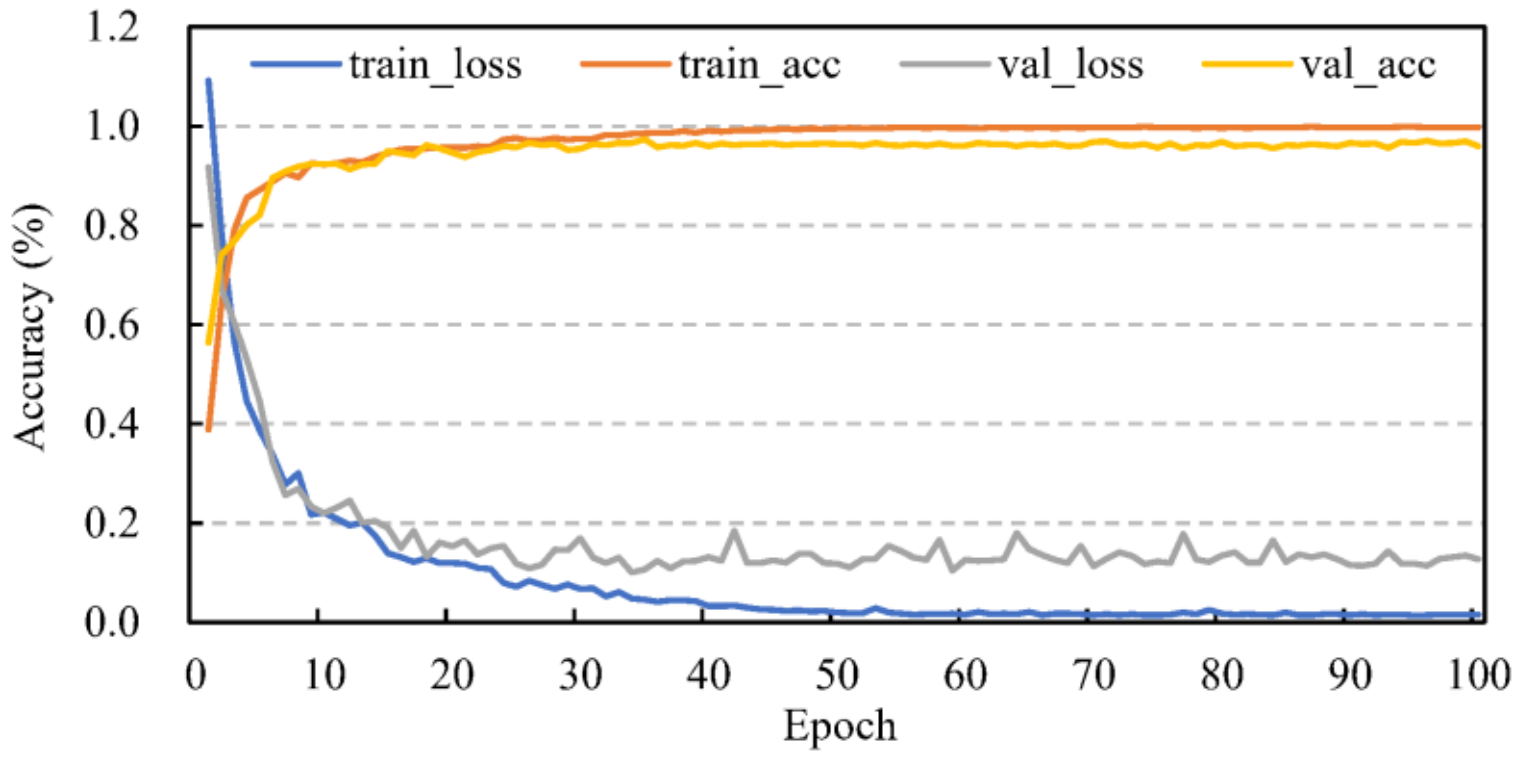

Figure 7. Accuracy and loss curves of PMIN on the training and validation sets.

## 4.3 Analysis of quantifying results

Figure 8 and Table 3 present the confusion matrix and classification performance of PMIN for fish feeding intensity quantification, respectively. Overall, PMIN shows strong discriminative ability across all feeding intensity levels, with the most stable recognition achieved for the "None". Among all None samples, only two are misclassified as "Weak", and no sample is incorrectly classified as "Strong". Accordingly, this class achieves a recall of 99.14%, a precision of 100%, and an F1-Score of 99.57%, indicating that the model can accurately capture the discriminative characteristics of multimodal signals under the non-feeding state. From the distribution of misclassification results, the errors are mainly concentrated between the "Weak" and "Strong", and neither of them is misclassified as "None". Given that fish feeding behavior and multimodal responses inherently exhibit continuity and gradual transitions, this confusion pattern is largely consistent with the category transition characteristics observed in the actual feeding process. A further examination of the class-wise evaluation metrics shows that PMIN also maintains relatively balanced performance on the two feeding-related classes. Compared with the "Weak", the "Strong" has slightly higher precision but slightly lower recall, suggesting that the model is relatively more conservative when identifying the "Strong". Overall, PMIN can accurately distinguish between non-feeding and feeding states while maintaining good discriminative ability between adjacent feeding intensity levels, which verifies its effectiveness for fish feeding intensity quantification.

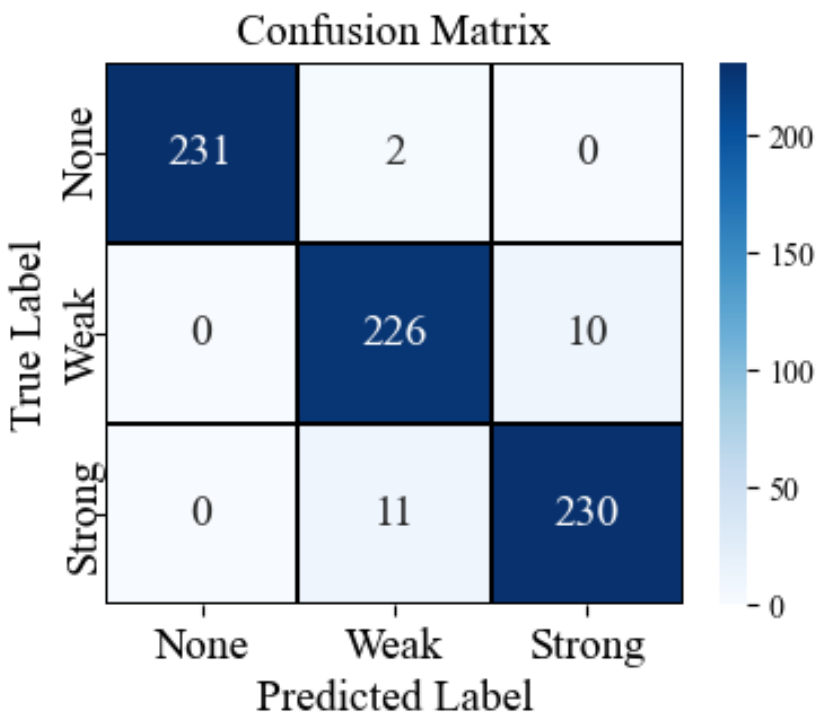

Figure 8. Confusion matrix of PMIN in quantifying fish feeding intensity.

Table 3. Quantification results of PMIN for different fish feeding intensities.

| Feeding intensity | Accuracy (%) | Precision (%) | Recall (%) | F1-Score (%) |
|---|---|---|---|---|
| None | 99.14 | 100 | 99.14 | 99.57 |
| Weak | 95.76 | 94.56 | 95.76 | 95.16 |
| Strong | 95.44 | 95.83 | 95.44 | 95.63 |

## 4.4 Comparison of different modal fusions

To verify the effectiveness of multimodal fusion for fish feeding intensity quantification, this study further compares model performance under unimodal, bimodal, and trimodal input settings, and the results are shown in Table 4. To ensure a fair comparison, UniRepLKNet is adopted for feature extraction in all modality settings, and the proposed DAIM-2 module is used for feature interaction under the bimodal setting.

Table 4. Performance comparison between single-modal and multimodal fusion models.

| Image | Audio | Water wave | Accuracy (%) | Precision (%) | Recall (%) | F1-Score (%) |
|---|---|---|---|---|---|---|
| √ | × | × | 82.39 | 83.08 | 82.41 | 82.74 |
| × | √ | × | 59.86 | 60.17 | 59.73 | 59.95 |
| × | × | √ | 66.34 | 66.25 | 66.32 | 66.28 |
| √ | √ | × | 91.13 | 91.12 | 91.17 | 91.15 |
| √ | × | √ | 92.39 | 92.36 | 92.43 | 92.40 |
| × | √ | √ | 83.80 | 84.39 | 83.88 | 84.13 |
| √ | √ | √ | **96.76** | **96.80** | **96.78** | **96.79** |

Note: √ indicates that this modality is used, × indicates that it is not used.

As shown in Table 4, multimodal fusion consistently outperforms unimodal methods in overall performance. This result indicates that fish feeding intensity has a clear dependence on multi-source information, and relying on a single modality is insufficient for comprehensive and robust representation. Among the unimodal models, the image modality achieves the best performance, with an accuracy of 82.39%, which is markedly higher than that of the water-wave modality at 66.34% and the audio modality at 59.86%. This finding suggests that visual information can reflect fish feeding behavior more directly and therefore provides stronger discriminative capability. Under the bimodal settings, the fusion of audio and water-wave signals increases the amount of available information to some extent, but the performance gain remains limited because the correlation between these two modalities is relatively weak. In contrast, when the image modality is fused with audio or water-wave data, the model performance improves substantially, with accuracy rising to 91.13% and 92.39%, respectively. This result indicates that the image modality forms a more effective complementary relationship with the other modalities and thereby enhances the representation capability of the model. Furthermore, the combination of image, audio, and water-wave modalities achieves the best performance, which further confirms the necessity and effectiveness of multimodal collaborative modeling. The image modality provides rich behavioral and scene information, whereas the audio and water-wave modalities supplement dynamic response and environmental interaction cues. These three modalities therefore form effective information complementarity and jointly improve the model's ability to quantify fish feeding intensity.

## 4.5 Ablation study

To evaluate the effects of the proposed ARPM and AER on the performance of PMIN, this study designs a detailed ablation experiment that includes a baseline strategy without improvement, single-improvement strategies, and a combined-improvement strategy, and the results are shown in Table 5. Specifically, the baseline strategy without improvement adopts Concat for multimodal feature fusion (Du et al., 2024). The single-improvement strategies examine the performance changes when only ARPM or only AER is introduced. The combined-improvement strategy integrates both ARPM and AER simultaneously.

Table 5. Ablation experiment results of the PMIN.

| ARPM | Primary modality | AER | Accuracy (%) | Precision (%) | Recall (%) | F1-Score (%) |
|---|---|---|---|---|---|---|
| × | × | × | 87.75 | 87.65 | 87.79 | 87.72 |
| √ | Image | × | 94.79 | 95.05 | 94.85 | 94.95 |
| √ | Audio | × | 90.85 | 91.04 | 90.89 | 90.96 |
| √ | Wave | × | 92.68 | 93.54 | 92.77 | 93.15 |
| × | × | √ | 89.01 | 89.43 | 89.09 | 89.26 |
| √ | All | √ | **96.76** | **96.80** | **96.78** | **96.79** |

Note: √ indicates that the corresponding improvement strategy is adopted, while × indicates that it is not adopted.

As shown in Table 5, both proposed improvement strategies contribute positively to the performance of PMIN. Without the introduced improvement modules, the baseline model achieves an accuracy of only 87.75%. After introducing AER, the accuracy increases to 89.01%, indicating that this strategy can alleviate decision inconsistency across modalities to a certain extent, although the overall gain remains limited. In contrast, introducing ARPM alone leads to a more substantial performance improvement. When image, audio, and water-wave modalities are respectively treated as the primary modality, the model accuracy increases by 7.04%, 3.1%, and 4.93%. This difference in improvement further suggests that, in fish feeding intensity quantification, the visual modality has stronger discriminative capability and more stable representations, making it more suitable as the dominant branch for guiding cross-modal feature enhancement. By comparison, the water wave modality provides certain auxiliary representational value, whereas the audio modality offers relatively weaker support when used as the primary modality. Furthermore, the complete model that simultaneously incorporates ARPM and AER achieves the best results on all evaluation metrics. This finding indicates that the two modules are functionally complementary rather than being a simple additive combination. ARPM mainly improves feature representation through progressive cross-modal interaction, whereas AER mitigates decision conflicts by modeling the confidence of modality-specific outputs. Since they operate at the feature learning stage and the decision fusion stage, respectively, they jointly improve the quantification performance of the model.

## 4.6 Effectiveness of the unified feature extraction framework

To verify the effectiveness of the unified feature extraction framework in fish feeding intensity quantification, this study further compares multiple configuration schemes, including homogeneous models in which all three modalities use the same feature extractor and heterogeneous models in which different modalities use different feature extractors. The results are shown in Tables 6 and 7. The adopted feature extractors include EfficientNet V2-S (Tan & Le, 2021), MobileNet V4-Conv-S (Qin et al., 2025), ConvNeXt V2-A (Woo et al., 2023), RepLKNet (Ding et al., 2022), UniRepLKNet (Ding et al., 2024), OverLoCK-XT (Lou & Yu, 2025), LSNet-T (A. Wang et al., 2025), ResNet18 (He et al., 2016), and Vision Transformer (Dosovitskiy et al., 2021).

Table 6. Performance comparison of homogeneous models.

| Feature extractors | Accuracy (%) | Precision (%) | Recall (%) | F1-Score (%) | Params (M) | FLOPs (G) |
|---|---|---|---|---|---|---|
| EfficientNet V2-S | 89.58 | 90.89 | 89.60 | 89.75 | 84.58 | 9.57 |
| MobileNet V4-Conv-S | 83.94 | 84.50 | 83.95 | 84.23 | 17.94 | 3.77 |
| ConvNeXt V2-P | 92.96 | 93.00 | 93.00 | 93.00 | 29.02 | 4.09 |
| LSNet-T | 83.52 | 88.35 | 83.46 | 82.87 | 35.04 | **0.96** |
| RepLKNet | 93.52 | 93.56 | 93.57 | 93.56 | 35.50 | 5.65 |
| OverLoCK-XT | 94.51 | 94.71 | 94.54 | 94.63 | 50.21 | 8.2 |
| ResNet18 | 81.69 | 81.64 | 81.81 | 81.71 | 36.90 | 5.36 |
| ViT | 91.27 | 91.25 | 91.31 | 91.28 | 25.23 | 4.48 |
| PMIN (Ours) | **96.76** | **96.80** | **96.78** | **96.79** | **14.39** | 1.88 |

Table 7. Performance comparison of heterogeneous models.

| Image | Audio | Wave | Accuracy (%) | Precision (%) | Recall (%) | F1-Score (%) | Params (M) | FLOPs (G) |
|---|---|---|---|---|---|---|---|---|
| URNet | EFNet | MNNet | 92.96 | 93.00 | 93.00 | 93.00 | 37.72 | 5.04 |
| RCNet | URNet | CNext | 91.97 | 92.03 | 92.11 | 92.07 | 18.98 | 3.59 |
| CNext | RSNet | URNet | 90.42 | 90.42 | 90.44 | 90.43 | **17.88** | 3.41 |
| LSNet | RCNet | EFNet | 89.58 | 89.55 | 90.71 | 90.13 | 45.20 | 5.21 |
| URNet | LSNet | ViT | 87.18 | 87.22 | 87.72 | 87.25 | 22.27 | 2.40 |
| URNet | MNNet | LSNet | 94.37 | 94.41 | 94.49 | 94.45 | 18.58 | **2.14** |
| ViT | RCNet | URNet | **95.92** | **95.94** | **95.95** | **95.95** | 21.23 | 3.87 |
| CNext | ViT | RSNet | 76.34 | 76.75 | 85.88 | 81.06 | 21.49 | 4.23 |
| URNet | MNNet | ViT | 95.63 | 95.85 | 95.62 | 95.61 | 19.89 | 4.29 |
| MNNet | LSNet | CNext | 86.93 | 86.27 | 86.04 | 86.03 | 20.03 | 2.74 |

Note: MobileNet V4-Conv-S (MNNet); ConvNeXt V2 (CNext); RepConvNet (RCNet); UniRepLKNet (URNet); ResNet18 (RSNet); Vision Transformer (ViT); EfficientNet V2-S (EFNet).

The experimental results show that the constructed unified feature extraction framework achieves the best overall performance. PMIN reaches 96.76%, 96.80%, 96.78%, and 96.79% in Accuracy, Precision, Recall, and F1-Score, respectively, which are consistently higher than those of all comparison models. This demonstrates that the proposed framework effectively improves the discriminative capability for fish feeding intensity quantification. The comparison results of homogeneous models show that although some backbone networks have achieved competitive results, there are still obvious limitations. For example, OverLoCK-XT achieves an accuracy of 94.51%, but its parameter counts and FLOPs are as high as 50.21M and 8.2G, respectively. Although LSNet-T and MobileNet V4-Conv-S provide certain advantages in lightweight design, their classification performance remains relatively weak.

In addition, the performance of heterogeneous models fluctuates substantially. For instance, when ViT, RCNet, and URNet are adopted for the image, audio, and water-wave modalities, respectively, the model achieves an accuracy of 95.92%. In contrast, when CNext, ViT, and RSNet are combined as heterogeneous extractors, the accuracy drops sharply to 76.34%. This instability essentially reflects the strong dependence of multimodal fish feeding intensity quantification on representational consistency. Image, audio, and water-wave signals differ markedly in information form, statistical distribution, and noise characteristics. Although different feature extractors may strengthen unimodal representation ability, they are also more likely to introduce inconsistency across modal representation spaces, which increases the difficulty of mining complementary information across modalities. In addition, from the perspective of efficiency, PMIN also shows a favorable balance between performance and complexity. While achieving higher classification accuracy, it requires only 14.39M parameters and 1.88G FLOPs, which is better than some comparison models that reach relatively high accuracy at a much higher computational cost. This advantage is particularly important for edge deployment and real-time monitoring in practical aquaculture scenarios, where fish feeding intensity quantification systems usually need to balance accuracy, inference speed, and hardware resource consumption.

## 4.7 Comparison of feature fusion methods

To verify the superiority of ARPM in multimodal feature fusion, this study further compares multiple feature fusion methods, including Concat (fusion through concatenation), Mul (fusion through element-wise multiplication), Avg (fusion by averaging weighted features), AW (fusion through adaptive weighting), and ARPM with the image as the primary modality. The results are shown in Table 8. To ensure a fair comparison, all fused features are uniformly input into a fully connected layer for final prediction.

Table 8. Performance comparison of different feature fusion methods.

| Methods | Accuracy (%) | Precision (%) | Recall (%) | F1-Score (%) | Params (M) |
|---|---|---|---|---|---|
| Concat | 87.75 | 87.65 | 87.79 | 87.72 | 13.05 |
| Mul | 89.01 | 89.09 | 89.43 | 89.26 | 13.01 |
| Avg | 83.38 | 84.03 | 83.40 | 83.71 | 13.01 |
| AW | 93.10 | 93.65 | 93.10 | 93.38 | 13.01 |
| ARPM | **94.79** | **95.05** | **94.85** | **94.95** | 13.45 |

Overall, ARPM achieves the best performance on all evaluation metrics. Specifically, the Accuracy, Precision, Recall, and F1-Score of ARPM reach 94.79%, 95.05%, 94.85%, and 94.95%, respectively, all of which are higher than those of the other comparison methods. In contrast, the three basic fusion methods, namely Concat, Mul, and Avg, show relatively weak overall performance, indicating that they are insufficient to effectively model the complex cross-modal dependencies involved in fish feeding intensity quantification. Avg yields the worst performance, suggesting that when different modalities exhibit inconsistent responses, direct smoothing-based fusion may weaken the discriminative power of critical information. In addition, AW achieves an accuracy of 93.10%, which indicates that adaptive weighting is more effective than basic fusion strategies and can alleviate the imbalance of modality contributions to some extent. However, its performance remains lower than that of ARPM, which suggests that static or shallow weighting mechanisms are still insufficient to capture deeper complementary relationships across modalities. By contrast, ARPM progressively enhances the representation of the primary modality under the guidance of auxiliary modality information, thereby forming more discriminative multimodal feature representations. It is worth noting that, compared with the other methods, ARPM increases the number of parameters by only 0.40M to 0.44M, while improving accuracy by 1.69% to 11.41%. This result indicates that its performance gain mainly comes from the proposed progressive multimodal interaction mechanism itself rather than from a simple expansion of model size, which demonstrates a favorable balance between model complexity and fusion performance.

## 4.8 Comparison of decision fusion methods

To verify the effectiveness of the AER in the result fusion stage for fish feeding intensity quantification, this study further compares multiple decision fusion methods, including Majority Voting (MV), Probability Averaging (PA), Learning-based Fusion (LF), Dempster-Shafer evidence theory (DS) (K. Zhao et al., 2022), and AER. The results are shown in Table 9. To ensure a fair comparison, all methods adopt the proposed ARPM as the feature fusion module.

Table 9. Performance comparison of different decision fusion methods.

| Methods | Accuracy (%) | Precision (%) | Recall (%) | F1-Score (%) | Inference speed (s) |
|---|---|---|---|---|---|
| MV | 82.96 | 83.62 | 82.97 | 83.29 | **0.00480** |
| PA | 90.84 | 91.04 | 90.89 | 90.96 | 0.00503 |
| LF | 94.36 | 94.44 | 94.35 | 94.36 | 0.00762 |
| DS | 91.83 | 92.44 | 91.93 | 92.18 | 0.00568 |
| AER | **96.76** | **96.80** | **96.78** | **96.79** | 0.00485 |

As shown in Table 9, the AER achieves the best performance among all compared methods, indicating that explicitly modeling evidence reliability and consistency at the decision level is more suitable for fish feeding intensity quantification than conventional rule-based or learning-based fusion strategies. Specifically, AER reaches 96.76%, 96.80%, 96.78%, and 96.79% in Accuracy, Precision, Recall, and F1-Score, respectively, all of which are higher than those of MV, PA, LF, and DS. By comparison, MV yields the weakest performance, suggesting that a fusion strategy based only on majority opinion is insufficient for tasks with inconsistent multimodal responses. PA averages the probability outputs from different sources and can therefore integrate multi-source information to some extent, leading to better performance than MV. LF further learns from the decision outputs of different modalities and strengthens multimodal decision integration, but it also introduces additional inference cost. In addition, DS also achieves relatively good results on several evaluation metrics, which indicates its advantage in handling uncertain information. However, the overall performance of AER still surpasses that of DS, suggesting that further incorporating weight and reliability modeling into the evidence fusion process helps more effectively alleviate evidence conflict and improve the final decision quality. Overall, the AER makes better use of the complementary information and confidence differences among multimodal outputs, thereby achieving superior decision fusion performance.

## 4.9 Application in real-world scenarios

To further evaluate the feeding intensity quantification capability of PMIN in a real aquaculture environment, this study conducts testing on an unseen dataset that is not involved in training. This dataset contains continuously collected video, audio, and water-wave signals with a total duration of 240 second. During data preprocessing, a sliding-window strategy is adopted, in which 1 second segments are used as model inputs with a step size of 0.5 second. Figure 9 shows a comparison between the model's quantization results and the ground truth labels.

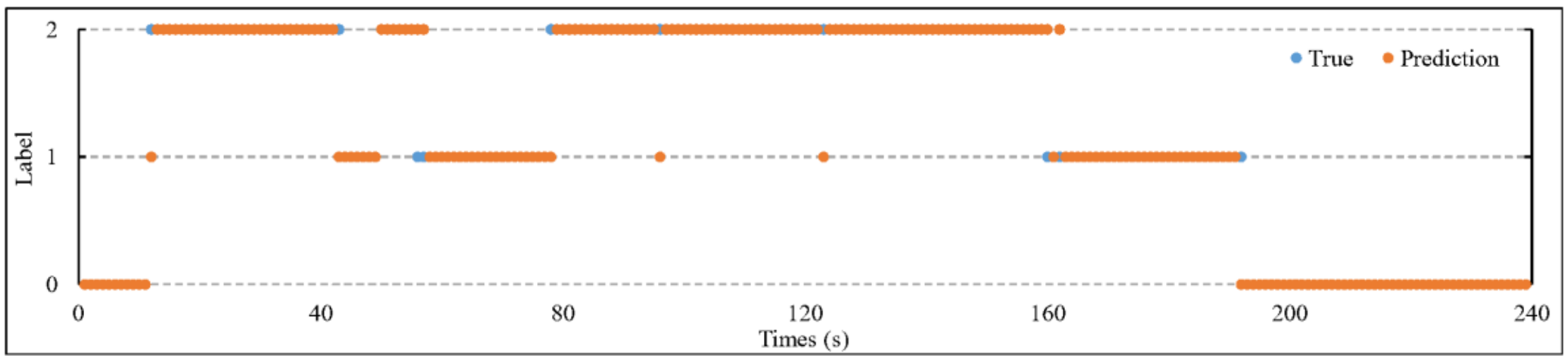

Figure 9. Comparison of the model’s predictions with true results.

Overall, the quantification results remain highly consistent with the temporal evolution of the true feeding intensity. Among the 239 sliding-window samples, the model correctly identifies 229, yielding an overall accuracy of 95.82%. In the early stage, the model can stably recognize the non-feeding state. As feeding proceeds, the quantification results shift in a timely manner to higher feeding intensity levels and remain consistent with the ground-truth labels over relatively long intervals. Similar behavior is also observed in the middle and later stages, indicating that PMIN not only achieves high recognition accuracy at the individual sample level, but also shows good temporal continuity and stability. This property is particularly important for practical feeding monitoring, because real-world management is more concerned with the sustained evolution of feeding states than with instantaneous predictions for a single window. Although the overall consistency is high, a small number of local deviations can still be observed in the figure, mainly around state transition regions and short weak-feeding segments. This phenomenon is reasonable in real scenarios, because the multimodal responses of adjacent feeding states are often highly similar, and the sliding-window mechanism may also introduce a smoothing effect on short-term behavioral changes. Therefore, these errors are more likely to arise from the ambiguity of state boundaries rather than from any systematic failure of the model. This real-scene experiment demonstrates that PMIN is suitable for real-time feeding intensity quantification in practical aquaculture systems and can provide effective support for precision feeding management.

# 5 Discussion

(1) The rationality of the unified feature extraction framework: Although the proposed unified feature extraction framework achieves strong performance in fish feeding intensity quantification, this does not necessarily imply that it is the optimal solution. In principle, selecting more suitable feature extractors for different modalities may lead to a better heterogeneous combination. This possibility is particularly relevant for image, audio, and water-wave signals, whose statistical properties, signal structures, and discriminative cues differ substantially. Feature extractors designed for a single modality may therefore offer stronger local representation capability. However, heterogeneous configurations usually involve a very large search space. When the number of candidate feature extractors is large, the possible cross-modal combinations grow rapidly, and the time cost and computational burden of model training, parameter tuning, and repeated validation also increase substantially. Under such conditions, it is unrealistic to evaluate all possible combinations exhaustively. Limited by the experimental schedule and available computational resources, this study cannot cover all heterogeneous configurations and instead reports only several representative combinations. Even so, the unified feature extraction framework still has clear value. On the one hand, it provides a modeling scheme with structural consistency, controllable implementation, and good interpretability for multimodal fusion, thereby reducing the interference caused by encoder differences. On the other hand, the framework already shows strong overall performance in the present task, which suggests that it achieves a reasonable balance among performance, complexity, and implementation controllability.

(2) The necessity of decision fusion: Existing multimodal studies usually perform final prediction directly through a fully connected layer after feature fusion, while an independent decision fusion stage is less frequently designed. Meanwhile, the results in Table 8 show that the model can still achieve good performance even without introducing decision fusion. Therefore, whether it is necessary to add an extra decision fusion stage beyond feature fusion is a question worth discussing. The main reason for introducing the AER for decision-level fusion in this study is that, in fish feeding intensity quantification, different modalities do not respond to the same state in a fully consistent manner, and their information quality and discriminative reliability may vary with the scene. As a result, relying only on fused features for one-shot prediction is often insufficient to fully exploit the uncertainty information and complementary decision cues contained in the outputs of different modalities. Although Table~8 indicates that the feature interaction module alone already yields strong results, the further introduction of AER still leads to a stable improvement in model performance. This finding suggests that decision fusion is not a simple repetition of feature fusion, but rather a subsequent complementary optimization step that further improves the reliability and consistency of the final output by explicitly modeling confidence differences and decision conflicts across modalities. Therefore, this paper argues that the introduction of decision fusion should be considered in conjunction with the specific task characteristics and experimental results.

# 6 Conclusion

This study proposes a Progressive Multimodal Interaction Network (PMIN) for fish feeding intensity quantification, aiming to alleviate response inconsistency and decision conflicts across modalities, thereby improving the accuracy and reliability of quantification results. Extensive experiments are conducted on a multimodal dataset containing image, audio, and water-wave signals to verify the effectiveness and superiority of the proposed method. Specifically, PMIN achieves 96.76%, 96.80%, 96.78%, and 96.79% in Accuracy, Precision, Recall, and F1-Score, respectively, consistently outperforming both homogeneous and heterogeneous comparison models and demonstrating strong overall performance. The experimental results further show that multimodal collaborative modeling plays an important role in fish feeding intensity quantification. Image, audio, and water-wave signals provide complementary information from the perspectives of behavioral representation, acoustic response, and environmental disturbance, respectively, thereby alleviating the limitations of a single modality in dynamic perception, noise suppression, and discriminative representation. The ablation studies and the comparison of different fusion strategies show that ARPM effectively enhances information interaction across modalities, while AER effectively handles uncertain information and decision conflicts, thereby improving the stability and reliability of model outputs. Real-scene application experiments further verify the potential of the proposed method in automated feeding monitoring and precision feeding decision-making. In addition, this study discusses the rationality of the unified feature extraction framework and the necessity of decision fusion. Future work will focus on cross-scene generalization ability, low-cost deployment, and robustness under long-term continuous monitoring, so as to promote the practical application of multimodal methods in smart aquaculture.

## CRediT authorship contribution statement

Shulong Zhang: Conceptualization, Investigation, Visualization, Writing–original draft, Writing–review & editing. Mingyuan Yao: Writing–review & editing, Methodology. Jiayin Zhao: Writing–review & editing, Visualization. Daoliang Li: Writing–review & editing, Methodology. Yingyi Chen: Writing–review & editing, Methodology. Haihua Wang: Methodology, Funding acquisition, Supervision, Writing–review & editing.

## Acknowledgments

This work was supported by National Key Research and Development Program of China (2023YFD2400400, 2023YFD2400401), R&D of Key Technologies and Equipment for Aquaponics

Intelligent Factory (CSTB2022TIAD-ZXX0053) and Mandarin Fish Factory Farming Service Project (202305510811525).

### Conflict of interest statement

The authors declare that there are no conflicts of interest.

### Data availability statement

The dataset is available at: https://huggingface.co/datasets/ShulongZhang/Multimodal_Fish_Feeding_Intensity.